\documentclass{jmlr}% new name PMLR (Proceedings of Machine Learning)

\usepackage{longtable}% for long tables

\usepackage{booktabs}
\usepackage[load-configurations=version-1]{siunitx} % newer version
 \usepackage{microtype}

\makeatletter
\def\set@curr@file#1{\def\@curr@file{#1}} %temp workaround for 2019 latex release
\makeatother

\usepackage{graphicx}
\usepackage{fullpage}
\usepackage{multicol}
\usepackage{multirow}
\usepackage{amssymb}
\usepackage{amsmath}
\usepackage{enumerate}
\usepackage{comment}
\usepackage{natbib}
\usepackage{hyperref}
\definecolor{gray}{rgb}{0.5,0.5,0}
\definecolor{green}{rgb}{0,0.42,0}
\newcommand*{\cellalign}[2]{\multicolumn{1}{#1}{#2}}

\title[BEDS-Bench: Behavior of EHR-models under Distributional Shift--A Benchmark]{
BEDS-Bench: Behavior of EHR-models under \\Distributional Shift--A Benchmark
}
\author{\Name{Anand Avati}
      \Email{avati@cs.stanford.edu}\\ 
      \addr Stanford University 
      \AND
      \Name{Martin Seneviratne}
      \Email{martsen@google.com}\\ 
      \addr Google
      \AND
      \Name{Emily Xue}
      \Email{yuanxue@google.com}\\
      \addr Google
      \AND
      \Name{Zhen Xu}
      \Email{zhenxu@google.com} \\
      \addr Google
      \AND
      \Name{Balaji Lakshminarayanan*}
      \Email{balajiln@google.com} \\
      \addr Google
      \AND
      \Name{Andrew M. Dai*}
      \Email{adai@google.com} \\
      \addr Google
      }
\def\beds{BEDS-Bench}
\def\bedS{BEDS-Bench }

\begin{document}

\maketitle

\begin{abstract}
  %Summary of the article.  Be sure to highlight how the workcontributes to our understanding of machine learning and healthcare.

%\section*{Abstract} 
Machine learning (ML) has recently demonstrated impressive progress in predictive accuracy across a wide array of tasks. Most ML approaches focus on generalization performance on unseen data that are ``similar'' to the training data (a.k.a. \emph{In-Distribution}, or IND). However, real world applications and deployments of ML rarely enjoy the comfort of encountering examples that are always IND. In such situations, most ML models commonly display erratic behavior on \emph{Out-of-Distribution} (OOD) examples, such as assigning high confidence to wrong predictions, or vice-versa. Implications of such unusual model behavior are further exacerbated in the healthcare setting, where patient health can potentially be put at risk. It is crucial to study the behavior and robustness properties of models under distributional shift, understand common failure modes, and take mitigation steps before the model is deployed. Having a benchmark that shines light upon these aspects of a model is a first and necessary step in addressing the issue. Recent work and interest in increasing model robustness in OOD settings have focused more on image modality, both in terms of methods as well as benchmarks, while the Electronic Health Record (EHR) modality is still largely under-explored. We aim to bridge this gap by releasing BEDS-Bench, a benchmark for quantifying the behavior of ML models over EHR data under OOD settings. We use two open access, de-identified EHR datasets to construct several OOD data settings to run tests on. The benchmark exercises several clinical prediction tasks, OOD data settings, and measures relevant metrics that characterize crucial aspects of a model's OOD behavior. We evaluate several learning algorithms under \bedS and find that all of them show poor generalization performance under distributional shift in general. Our results highlight the need and the potential to improve robustness of EHR models under distributional shift, and \bedS provides one way to measure progress towards that goal.
Code to reproduce the results in this paper and evaluate new algorithms against \bedS is made available at \url{https://github.com/Google-Health/records-research/tree/master/beds-bench}.
\end{abstract}

%\textbf{Keywords:} Out-of-Distribution, Electronic Health Records, Benchmark, Uncertainty Estimation.

\section{Introduction} \label{sec:intro}

\begin{figure}
    \centering
    \includegraphics[scale=0.15]{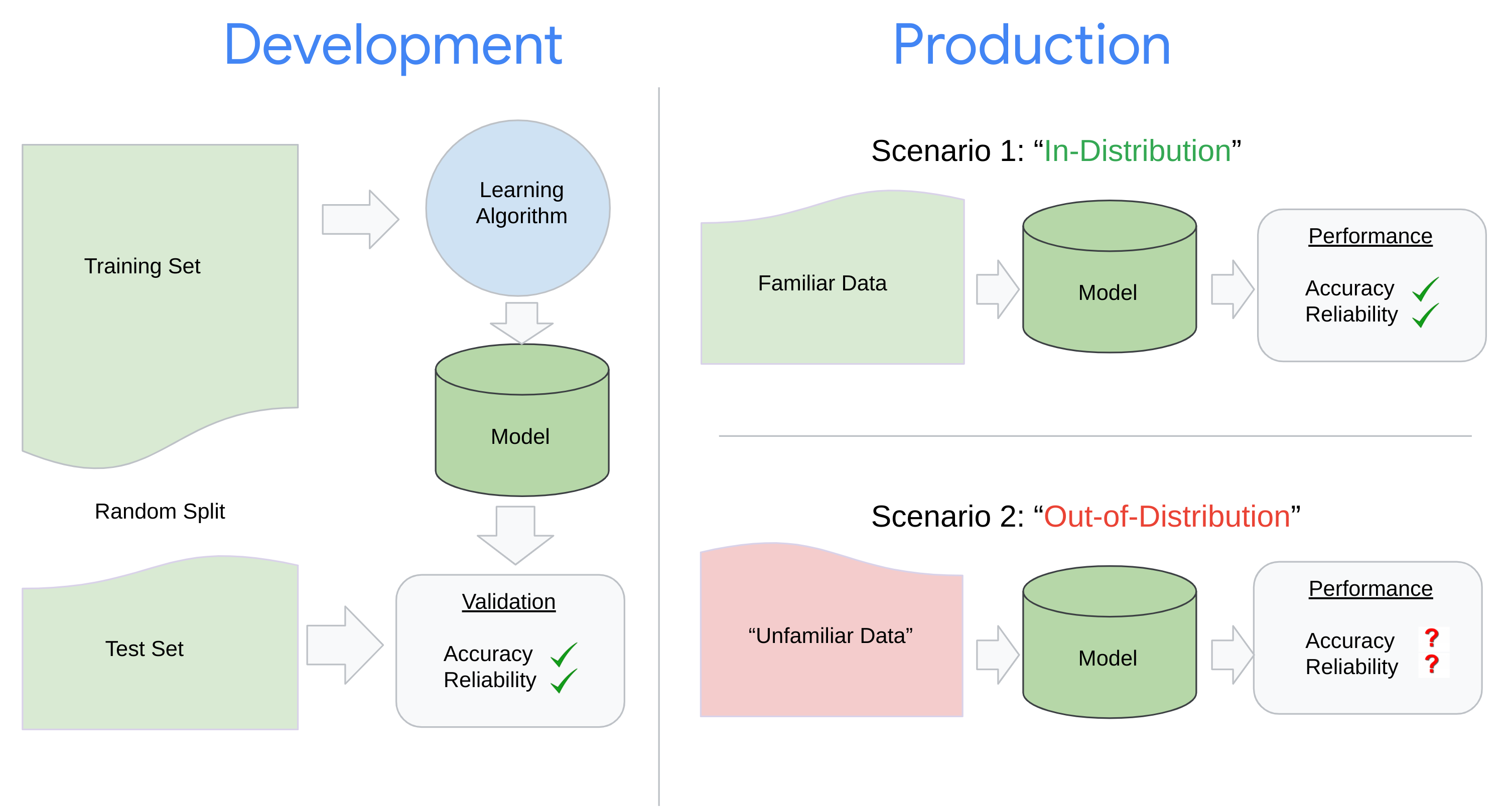}
    \caption{Illustration of a scenario where a model can encounter data that is very dissimilar to the training data distribution.}
    \label{fig:devprod}
\end{figure}

%\subsubsection*{What is the problem?}

Machine Learning models are typically validated on test sets that are similar to the training set. A common assumption in statistical learning theory is that all examples (both train and test) are drawn independently and identically from the same data distribution (IID). Though the IID assumption is a strong one, in practice it is hard to ascertain if it is always being met. When an ML model is deployed in a real world setting, the likelihood of encountering OOD inputs is far higher. In situations when an ML model is presented with OOD inputs, its behavior can be hard to describe theoretically, and tends to be unknown practically (Figure \ref{fig:devprod}). The first step in fixing the behavior of models in OOD settings is to measure and quantify it with benchmarks. Benchmarks and datasets paint a target for the research community to focus and align on, thereby catalyzing the progress of the field \citep{imagenet_cvpr09,uci}. They  also serve a crucial role as an objective measure of progress towards that goal. Yet, there is a lack of good benchmarks for studying the behavior of models on EHR data under OOD settings, which our work attempts to address.

%\subsubsection*{Why is it interesting and important?}

Studying the behavior of EHR models under distributional shift is more than just a purely academic endeavour \citep{nestor2019feature}. There are numerous real world situations where a model may encounter patients who are systematically different from the training data for legitimate reasons. Some specific examples where the train and test distributions may differ include: 
\begin{itemize}

    \item \textbf{Changes in the patient population:} The demographics of a patient population may change over time due to gentrification of neighborhoods around a health system, maturing public health policies, global population dynamics etc. Consider for example the rising proportion of females in the Veterans Affairs agency. This may result in models encountering patients from a different distribution than the historical data on which the model was trained.

     \item \textbf{Changes in the practice of medicine:} The COVID-19 pandemic is an example of a dramatic shift in the field of medicine as a whole. It introduced major distributional shifts via changes in the patient population, but also changes in the practice of medicine, the therapies being used and the operational processes of the hospital (e.g. due to resource shortages).
     
    \item \textbf{Portability of models between health systems:} There is increased sharing of pre-trained EHR models between hospital sites, with vendors offering pre-built models \citep{TAN2020575} and academic consortia such as the Observational Health and Data Sciences Initiative (OHDSI) enabling model portability via common data standards. While this is excellent for broadening the impact of machine leaning and encouraging research reproducibility, it also increases the likelihood of training and deployment datasets being divergent due to differences in both populations and data formats.

\end{itemize}

% \begin{comment}
%     \item Global events (e.g. The Olympics) attract bursts of foreign visitors, some of whom may have encounters with the local health system. It is unlikely that ML models trained only on data in the local health system would have witnessed patients with diverse demographics.
% \end{comment}

When the behavior of an EHR model under distributional shift is unknown, there is a risk that predictions on OOD inputs might be wrong yet highly confident, thereby potentially increasing clinical risk for those patients. This is particularly important as EHR models start to be deployed in real world clinical settings \citep{translation}.  

%\subsubsection*{Why is it hard? (E.g., why do naive approaches fail?)}

While OOD benchmarks have been extremely impactful in the imaging domain, creating an analagous EHR benchmark is challenging. First, privacy concerns makes it hard to even get access to multiple large EHR datasets. In addition, EHR data is complex, heterogeneous and highly site-specific. This makes it difficult to harmonize multiple EHR datasets in order to perform cross-site experiments to evaluate OOD behaviour. Furthermore, while the benchmark tasks in the imaging domain is typically a classification problem with readily available labels, EHR tasks are often less straightforward. For example, defining a task involving EHR data necessarily involves nuanced data and temporal considerations, such as deciding a consistent \emph{prediction time} for all examples (e.g. predicting onset of diabetes is meaningful only when the disease is not yet diagnosed), choice of a suitable time window and data sources from which data is extracted for features (broad window makes for more accurate models, but reduces the population who have sufficient data to be applied upon), determining a suitable representation for the extracted sparse and heterogenous data (handling a mixture of real values, categorical values, ordinal values, timestamps, handwritten text, images, missing values etc.), assigning labels (e.g. how to accurately determine which patients actually have diabetes), among other challenges.

%\subsubsection*{Why hasn't it been solved before? (Or, what's wrong with previous proposed solutions? How does mine differ?)}

%\subsubsection*{What are the key components of my approach and results? Also include any specific limitations.}

\begin{figure}
\centering
\includegraphics[scale=0.15]{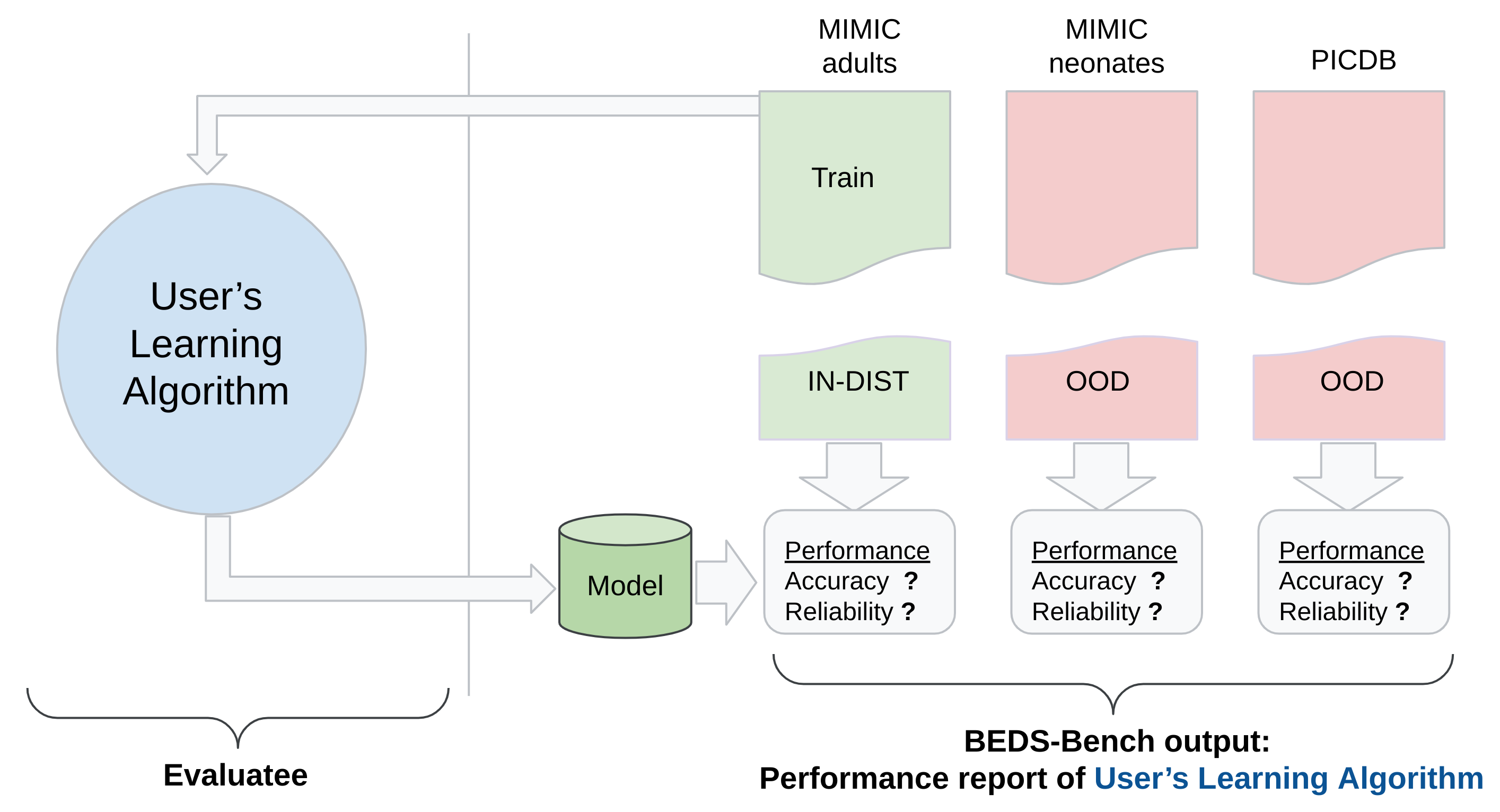}
\caption{\bedS setting}
\label{fig:bedsbench1}
\end{figure}

To this end, \bedS is a benchmark created using two open access de-identified EHR datasets. \bedS simulates OOD settings by creating intentionally dissimilar train and test sets, and measures several metrics around model performance in each of these settings (see Figure \ref{fig:bedsbench1}), for three common downstream classification tasks. The code for pre-processing and model evaluation is open-sourced and we hope that these benchmarks are a useful resource for the EHR community to develop more rigorous methods to characterize OOD behaviour. 

\vspace{4mm} 

\textbf{Summary of contributions.} We summarize our contributions below:
\begin{enumerate}%[i]
    \item We design an OOD benchmark on EHR data that includes suitable definitions of data partitions and splits, downstream tasks, and evaluation metrics.
    \item We harmonize two EHR data sources to enable cross-dataset experimentation.
    \item We evaluate several algorithms on this benchmark and report on their performance.
\end{enumerate}

The rest of the paper is organized as follows. In Section~\ref{sec:related} we give an overview of related work around robustness to distributional shift as well as various benchmark efforts. In Section~\ref{sec:beds} we describe the \bedS benchmark in detail. Sections~\ref{sec:experiments} and \ref{sec:discussion} describe the experiments and results, and we conclude with Section~\ref{sec:conclusion}.

\section{Related work} \label{sec:related}
\citet{nestor2019feature} used a timestamped version of the MIMIC-III dataset to demonstrate significant deterioration in model performance when EHR models were evaluated on data more recent than the training set. Their proposed mitigation strategy involved harmonizing features into clinical concept groupings. While pre-processing strategies can be effective, there is a complementary need for better model-based strategies for OOD detection and mitigation, motivating the present work.

%\begin{comment}
While there is a paucity of literature on EHR robustness evaluation, there has been some progress with the images modality such as the Imagenet-c dataset \citep{imagenetc}. Many image based OOD works utilize multiple datasets, such as MNIST \citep{lecun-mnisthandwrittendigit-2010}, ImageNet \citep{imagenet_cvpr09}, SVHN \citep{svhn}, etc. to conduct cross-dataset experiments to analyze model behavior \citep{deepgenknow}. There have also been methods developed to improve model calibration and robustness to OOD examples, though these works mostly focused their experiments and efficacy tests on images \citep{deepensembles, sngp}.
%\end{comment}

The most related work to ours is a recent paper that has evaluated several ML algorithms on their ability to detect OOD EHR inputs by assigning a higher uncertainty in their outputs \citep{trustissues}. Their focus is limited to OOD \emph{detection}, while \bedS takes a more holistic view of model behaviour under distributional shift (described in Section~\ref{sec:ideal}). We discuss additional challenges of OOD detection under class imbalance and certain choice of uncertainty metrics in Section~\ref{sec:discussion}.

%\begin{itemize}
%    \item Our cross-dataset test (MIMIC vs PICDB) has more contrast (different ages, country, ethnicity, medical practices) vs (MIMIC vs eICU - both within US and largely adult)
%    \item We have multiple downstream tasks (Mortality, LoS3+, LoS7+) whereas they have just mortality
%    \item Their ethnicity test includes only black vs white (ethically questionable?  Could lead to models being good at robust against just black vs white and not other races?). Asian and Hispanic is too few in the dataset.
%    \item Their methodology does not seem to have a sufficient (6hr) “gap” between X and Y which is a standard practice, which otherwise can result in “label leak”.
%\end{itemize}

\section{\beds} \label{sec:beds}

The \bedS tool is designed to generate a performance report of a learning algorithm regarding the behavior of  models trained by this algorithm under various types of distributional shift in the test data. The general approach taken by \bedS is to partition data in several ways into \emph{intentionally} dissimilar subsets in order to artificially simulate  IND vs OOD settings. Models are trained on the train split of a certain subset for one of the standardized tasks, and tested on the test splits of all the subsets while measuring relevant metrics. The test split corresponding to the subset from which the model was trained is considered IND, while test splits from the other subsets in the partition are considered OOD. Figure \ref{fig:bedsbench1} describes the workflow in one particular setting. This procedure is repeated by cycling through every subset in every partition to be the IND, while the other subsets in the partition are considered OOD.

In the rest of this section we describe what is an ideal model's behavior under OOD data, the details of the methodology of \beds\ including descriptions of the datasets used, partitions created, tasks for which models are trained for, and metrics measured on the test sets.

\subsection{Ideal Model Behavior} \label{sec:ideal}

Before designing a benchmark, it is crucial to first define what we consider is the ideal behavior of a model. The tests and metrics of the benchmark need to then be chosen to shine light on these aspects of the model and enable objective comparison across multiple algorithms. The following notation of the ideal model behavior informs the design of \beds:
\begin{itemize}
    \item Generalization: When a model is tested on a distribution that is different from the one it was trained on, it is possible, and understandable, for the model performance to drop to some extent. A common generalization metric (in case of classification tasks) is the Area Under the Receiver-Operator Characteristic Curve (AUC) which measures the ability of a model to discriminate between two classes. The drop in generalization performance might likely be larger for test distributions that are ``farther'' from the train distributions. Yet, \emph{an ideal model should have at least a minimal level of generalization robustness to OOD data, such as not performing worse than random guessing (i.e. maintain AUC $\ge$ 0.5)}.

    \item Calibration: Calibration refers to the property that probabilities output by a model are agree with the observed empirical frequency of events. For example, among all days which had a rain forecast probability of 80\%, approximately 8 out of 10 days should observe rain in the long run. Calibration is a property that is orthogonal to discrimination, and hypothetically it is possible to have models with any mix of levels of calibration and discrimination. \emph{An ideal model is not only well-calibrated in its predictions on IND data, but also on OOD data, especially when generalization on OOD data has worsened.}

    \item Confidence: Closely related to the notion of calibration is \emph{confidence}. Typically confidence of a prediction is measured with metrics such as predictive entropy, or predictive variance. The larger the entropy or variance, the lower the confidence of that prediction. \emph{If an ideal model's OOD generalization performance is lower than IND, then the confidence in the OOD predictions will be lower than in the IND predictions}.
    
    While we do measure the ability of a model to discriminate OOD vs IND inputs by assigning lower confidence scores to OOD, we also emphasize that this test involves additional nuances that need to be considered before interpreting the results. We discuss this further in Section~\ref{sec:discussion}.
\end{itemize}

\subsection{Datasets}

\begin{table}[]
    \centering
 \begin{tabular}{|l|l|l|}
\hline
     & \cellalign{c|}{MIMIC} & \cellalign{c|}{PICDB} \\
   \hline
    Center & Beth Israel Deaconess Medical Center & Children's Hospital of Zhejiang University\\
    & & School of Medicine \\
     \hline
     Duration & 2001 - 2012 & 2010 - 2018 \\
     \hline
     Patient Count & 46,520 & 12,881 \\
     \hline
     Age range & 0-1mo, 16-89yrs (obfuscated at 90) & 0-18yrs \\
     \hline
     Encounters & 58,977 & 13,450 \\
     \hline
     Diagnostics & ICD-9 (6986 unique) & ICD-10CN (1122 unique) \\
     \hline
     Medication & NDC (4211 unique) & NCCD (657 unique)\\
     \hline
     Chart Events & 330,712,484 Events (6646 types) & 2,278,979 Events (19 types) \\
     \hline
     Lab Events & 27,854,056 Events & 10,094,118 Events \\
      & (727 unique, 137 LOINC) & (822 unique, 118 LOINC) \\
      \hline
      Microbiology & 631,727 Events (94 unique) & 183,870 Events (43 unique) \\
      \hline
      Procedures & 240,096 ICD-9 (3833 unique) & N/A \\
       & 573,147 CPT (2019 unique) & \\
       \hline
       Notes & 2,083,180 Notes (15 types) & N/A \\
           \hline
\end{tabular}   
    \caption{Summary and comparison of the MIMIC-III and PICDB datasets.}
    \label{tab:mimicpicdb}
\end{table}

To develop the benchmark, we make use of two open access Intensive Care Unit (ICU) EHR datasets -  Medical Information Mart for Intensive Care III (MIMIC-III, or MIMIC) \citep{mimiciii}, and Paediatric Intensive Care database (PICDB) \citep{picdb}. Both the datasets are available for download from PhysioNet \citep{PhysioNet}.

The MIMIC dataset has data related to patients who were admitted to the ICU at Beth Israel Deaconess Medical Center between 2001 and 2012. The dataset covers 58,977 ICU stays of 46,520 patients. All the patients were either adults or neonates (newborn babies).

The PICDB dataset was collected at the Children's Hospital of Zhejiang University School of Medicine between 2010 and 2018. It covers a total of 13,450 ICU stays of 12,811 patients who were all minors (newborn up to 18 years of age).

An overview and comparison of the two datasets, including types of data present in each dataset and their encoding formats is presented in Table \ref{tab:mimicpicdb}. Both the datasets are represented as relational databases, with a comma separated value (CSV) formatted file per table.

\subsection{Harmonization}

The way \bedS works is by creating intentionally dissimilar subsets of data to simulate OOD settings. One natural setting is to consider model behavior when trained on data from MIMIC and tested on PICDB data, and vice versa. Conducting such cross-dataset experiments is quite common, and straight forward with image data. For harmonizing two images datasets, the main considerations are around matching the resolutions, channel count, bits per color etc. which are all quite easily handled. Yet, harmonizing two different sources of EHR data is a lot more involved, with careful considerations required in finding a common set of tables, vocabularies (to codify categorical data), units (to represent continuous data), representation of time, and other semantic reconciliations.

The broad strategy we follow in harmonizing the two datasets is to identify a subset of tables, columns, and rows which can potentially be matched up, and exclude the remaining.

In PICDB the diagnostic codes are coded in the Chinese Edition of the International Classification of Diseases, Tenth Revision (ICD-10CN), whereas MIMIC uses the International Classification of Diseases, Ninth Revision (ICD-9). We perform a one-to-many mapping from ICD-10CN to ICD-9 using the Unified Medical Language System (UMLS) database \citep{umls}.

For medication codes, we map both the data sources to the RXCUI coding. MIMIC medications are coded in the National Drug Code, which uniquely map to RXCUI. For PICDB we start with the textual descriptions of the medications and run them through the MedEx system to extract the RXCUI codes \citep{medex}. 

While the laboratory tests are coded with custom codes in both the datasets, some of the custom codes have an accompanying Logical Observation Identifiers Names and Codes (LOINC) code. We use the LOINC code as the common vocabulary and include only those rows for which the custom code has a corresponding LOINC code.

MIMIC has a very rich representation of vitals and chart events. PICDB on the other hand has a total of nineteen vital and chart event types. We use the event type groupings from the MIMIC-Extract project to map a subset of the MIMIC chart event codes to the corresponding PICDB chart event codes \citep{mimicextract}.

The Inputevents and Outputevents table record the total volumes of different types of fluids that enter and exit the patient during the stay. While MIMIC records both the volumes and the types of the fluids, PICDB only records the volumes (without an associated fluid type). From a medical perspective, while knowing the type of fluid is certainly useful, just knowing the volume of fluids going in and out of the patients is also informative in itself. Thus we exclude the fluid type codes from MIMIC and retain only the volume information for the purposes of harmonization.

Table \ref{tab:harmonize} in the Appendix summarizes the various code harmonization approaches that were applied.

\subsection{Data Processing}

In order to create a supervised learning dataset out of an EHR relational database, certain additional data processing steps are necessary. Each example in the supervised learning dataset corresponds to the data from one hospital admission. First, we exclude all hospital admissions that are shorter than 30 hours, and within those included, we use data up to the first 24 hours since admission. The additional 6 hours ``gap'' after the first 24 hours of data is common practice to avoid leaking of information of the label into the covariates \citep{mimicextract}. Further, we only include the first admission of a patient, and exclude admissions after the first discharge, if any. Finally the dataset is randomly divided into train and test splits (80\% train, 20\% test).

The set of resulting tables after applying the inclusion and exclusion criteria (including both train and test), with their row and column counts is summarized in Table \ref{tab:resultingdata} in the Appendix. This is the harmonized dataset using which the various experimental OOD settings are created.

\subsection{Data Splits}

\begin{table}[]
    \centering
    \begin{tabular}{|c|l|l|}
    \hline
    Partition & \cellalign{c|}{Slice} & \cellalign{c|}{Criteria} \\
    \hline
    \hline
    \multirow{3}{*}{Demographics} & MIMIC-adult & DB = ``MIMIC'' and AGE $>$ 15yr \\
    \cline{2-3} 
    & PICDB-paed & DB = ``PICDB'' \\
    \cline{2-3}
    & MIMIC-neonate & DB = ``MIMIC'' and AGE $<$ 1mo \\
    \hline
    \hline
    \multirow{2}{*}{Biological Sex}     & MIMIC-Female & DB = ``MIMIC'' and Gender = ``F'' \\
    \cline{2-3}
    & MIMIC-Male & DB = ``MIMIC'' and Gender $\ne$ ``F'' \\
    \hline
    \hline
     \multirow{5}{*}{Ageing}    & MIMIC-lt50 & DB = ``MIMIC' and 15yrs $<$ AGE $\le$ 50yrs \\
     \cline{2-3}
     & MIMIC-5060 & DB = ``MIMIC' and 50yrs $<$ AGE $\le$ 60yrs \\
     \cline{2-3}
     & MIMIC-6070 & DB = ``MIMIC' and 60yrs $<$ AGE $\le$ 70yrs \\
     \cline{2-3}
     & MIMIC-7080 & DB = ``MIMIC' and 70yrs $<$ AGE $\le$ 80yrs \\
     \cline{2-3}
     & MIMIC-gt80 & DB = ``MIMIC' and 80yrs $<$ AGE \\
     \hline
    \end{tabular}
    \caption{Data partitions and slices to construct IND vs OOD settings}
    \label{tab:partitions}
\end{table}

The benchmark creates three different partitions of the data, each partition having between two to five slices. Within each partition, the slices are completely non-overlapping, and are \emph{characteristically different} to varying degrees depending on the partition. The names and  definitions (inclusion criteria) of each of the slices of all the partitions are in Table \ref{tab:partitions}.

The Demographics partition has three slices - MIMIC-adult, MIMIC-neonate, and PICDB-paed. The differences in the slices in this partition are somewhat stark. Not only are the differences between paediatric (especially neonates) and adults particularly pronounced, the MIMIC vs PICDB slices present even more differences, including very distinct populations, health systems and accompanying treatment practices, etc. 

The Biological Sex partition separates the MIMIC dataset into Female and Male slices. Both the EHR datasets codify sex as binary and \beds follows the convention. This partition intends to highlight the model behavior under the extreme cases of shift in gender balance.

The Ageing partition slices the adults into different age bands, representing progressively older patients with each band. The age ranges in years used to define the bands are (15-50], (50-60], (60-70], (70-80] and (80,$\infty$).

It may be observed that some of the partitions have slices which are so blatantly dissimilar that sometimes it would be unreasonable to expect a model to ever generalize over to such a distinctly different dataset, or to even consider such generalization goals as clinically relevant.  Yet, we argue that these obviously-OOD settings are great examples of scenarios where any reasonably safe model would necessarily need to display some degree of robustness, and hence make for good tests to be included as part of an OOD benchmark suite.

We also note that the distribution of Race in the EHR datasets is quite skewed, with several races having too few examples to be sufficient to form a partition that includes slices for all races. After considerations of fairness and ethics, we look forward to finding additional EHR datasets that will allow us to construct a more inclusive race based partitioning.

\subsection{Supervised Learning Tasks}

\bedS includes three supervised learning tasks to evaluate algorithms on: In-Hospital Mortality (Mort), Remaining Length-of-Stay $>$ 3 days (LoS3+), and Remaining Length-of-Stay $>$ 7 days (LoS7+). All the three tasks are common canonical EHR tasks widely explored in the literature \cite{mimicextract, googleehr}, and framed as binary classification, with names suggestive of their labels. The Mort task has a label of 1 only if the patient passed away during the hospital stay of that example. Even if the patient passed away soon after discharge or during a follow-up admission, the label remains 0. The LoS3+ (or Los7+) task has a label of 1 only if the patient will end up having at least 3 (or 7) days worth of remaining time in their current stay.

The class balance varies significantly depending on the task and data slice. While mortality of MIMIC-neonate can be as low as 0.5\% (fortunately) on the one hand, the three day length of stay for the PICDB-paed slice is as high as 91.2\% on the other. The class balances for each of the three tasks on all the slices, along with the number of examples in each slice is listed in the Appendix (Table~\ref{tab:classbalance}).

\subsection{Metrics and Report} \label{sec:metrics}

The \bedS evaluates the performance of an algorithm in several test settings as measured by several metrics. For notation, let use denote the number of examples by $n \in \mathbb{N}$, $i \in \mathbb{N}$ as the example index number where $1 \le i \le n$, $y_i \in \{0, 1\}$ as the label (correct answer) of the $i^{th}$ example, and $\hat{y}_i \in [0,1]$ as the predicted probability by a model for the $i^{th}$ example.

\begin{itemize}
    \item \textbf{Task-AUC} - This metric is the Area Under the Receiver Operator Characteristic (ROC) curve (AUROC) measured in the context of the model predicting the downstream task label (Mort, LoS3+, LoS7+).
    \item \textbf{ECE} - Expected Calibration Error. To define ECE, we first divide the probability range [0,1] into $K$ equal non-overlapping intervals, each interval denoted $I_k, k \in [K]$. We also define $K$ corresponding bins $B_k, k \in [K]$ where each bin is the collection of example indices whose predicted probability falls in the interval $I_k$, i.e. $B_k = \{ i : \hat{y}_i \in I_k \}$. With this, the ECE is defined as $\text{ECE} = \frac{1}{n} \sum_{k=1}^K \left| \sum_{i \in B_k} \left( y_i - \hat{y}_i\right)\right|.$
    
    % \begin{align*}
    %     \text{ECE} &= \frac{1}{n} \sum_{k=1}^K \left| \sum_{i \in B_k} \left( y_i - \hat{y}_i\right)\right|.
    % \end{align*}
    
    \item \textbf{OOD-AUC} - This metric measures the ability of a model to assign higher confidence to IND test examples and lower confidence to OOD test examples. This metric requires both IND and OOD test sets for its calculation, whereas the previous two metrics are measured with only one test set (either IND or OOD) at a time. Confidence is typically considered to be the variance or entropy of the predicted Bernoulli distribution in case of a binary classification task. The AUC is measured with the label being set to 1 if the example is OOD (and 0 if IND), and the confidence measure is the score assigned to the example. The OOD-AUC will be high when OOD examples have higher variance or entropy than IND examples. Since both the variance and entropy of a Bernoulli distribution are similarly ordered (with $\hat{y}$=0.5 having the highest variance or entropy, and $\hat{y}$=0 or $\hat{y}$=1 having the lowest variance or entropy), the resulting OOD-AUC metrics with either choice will be the same.
\end{itemize}

These metrics are measured for each downstream task, on each data slice (IND) and other data slices within the same partition (OOD). The metrics are tabulated by algorithm, presenting the metrics of different algorithms in the same setting side by side.

\section{Experiments} \label{sec:experiments}

We evaluate seven algorithms on \bedS and analyze their performance: Logistic Regression (LogReg), Gaussian Process (GP) \citep{gpml}, Random Forest (RF)\citep{randomforest}, Mondrian Forest (MF)\citep{mondrianforest}, Multi Layer Perceptron (MLP), Bayesian Recurrent Neural Network (BRNN) \citep{brnnorig} with the same setup as \citep{brnn}, and Spectral-normalized Neural Gaussian Process (SNGP) \citep{sngp}. We use both the Scikit-Learn \citep{sklearn} and Tensorflow \citep{TF} software frameworks for the experiments, depending on the specific algorithm. Six of these models use a fixed length representation and one of them (BRNN) uses sequential embedding representation. The summary of models evaluated in this work is presented in Table \ref{tab:algos} in the Appendix.

The fixed length representation is calculated as an array of binary indicators for each of the possible codes that might appear in the training data. Age is represented in years, and volumes (inputevents and outputevents) are aggregated over the 24-hour period. The inputs are standardized  by column.

The sequential embedding representation creates an embedding for each categorical data and maintains the temporal ordering of all the codes. The data format of the generated representation matches that of \citep{tfseq}.

In all our experiments, within each partition, we randomly subsample (without replacement) the training examples to the size of the smallest training slice. This keeps all the training sets of equal size and makes is easy to compare metrics across different training slices.

\begin{table}[]
    \centering
    \small

\begin{tabular}{|l|l|c|c|c|c|c|c|c|}
\hline
\multicolumn{ 9 }{|c|}{ In Hospital Mortality  - AUC (larger values are better)} \\
\hline
Train & Test & LogReg & GP & RF & MF & MLP & BRNN & SNGP\\
\hline
\multirow{ 3 }{*}{{ Adult }}
& { Adult }
& \color{gray}{0.706}
& \color{gray}{0.703}
& \color{gray}{0.736}
& \color{gray}{0.650}
& \color{gray}{0.691}
& \color{gray}{ \textbf{0.784} }
& \color{gray}{0.710}
\\
\cline{2- 9 }
& { Neonate }
& \color{blue}{0.935}
& \color{blue}{ \textbf{0.946} }
& \color{black}{0.650}
& \color{orange}{0.122}
& \color{black}{0.836}
& \color{orange}{0.394}
& \color{blue}{0.878}
\\
\cline{2- 9 }
& { Paediatric }
& \color{orange}{0.459}
& \color{black}{ \textbf{0.685} }
& \color{black}{0.567}
& \color{black}{0.590}
& \color{orange}{0.307}
& \color{orange}{0.381}
& \color{orange}{0.528}
\\
\cline{2- 9 }
\hline
\hline
\multirow{ 3 }{*}{{ Neonate }}
& { Neonate }
& \color{gray}{0.790}
& \color{gray}{ \textbf{0.970} }
& \color{gray}{0.759}
& \color{gray}{0.890}
& \color{gray}{0.777}
& \color{gray}{0.762}
& \color{gray}{0.956}
\\
\cline{2- 9 }
& { Adult }
& \color{black}{0.562}
& \color{black}{ \textbf{0.594} }
& \color{black}{0.540}
& \color{black}{0.529}
& \color{orange}{0.440}
& \color{orange}{0.383}
& \color{orange}{0.459}
\\
\cline{2- 9 }
& { Paediatric }
& \color{orange}{0.555}
& \color{black}{0.610}
& \color{black}{0.577}
& \color{orange}{0.438}
& \color{orange}{0.398}
& \color{black}{0.602}
& \color{black}{ \textbf{0.615} }
\\
\cline{2- 9 }
\hline
\hline
\multirow{ 3 }{*}{{ Paediatric }}
& { Paediatric }
& \color{gray}{0.814}
& \color{gray}{0.813}
& \color{gray}{ \textbf{0.834} }
& \color{gray}{0.747}
& \color{gray}{0.796}
& \color{gray}{0.787}
& \color{gray}{0.799}
\\
\cline{2- 9 }
& { Adult }
& \color{orange}{0.488}
& \color{orange}{0.480}
& \color{orange}{ \textbf{0.500} }
& \color{orange}{0.492}
& \color{orange}{0.495}
& \color{orange}{0.426}
& \color{orange}{0.492}
\\
\cline{2- 9 }
& { Neonate }
& \color{black}{0.893}
& \color{black}{0.905}
& \color{blue}{0.944}
& \color{black}{0.702}
& \color{blue}{ \textbf{0.950} }
& \color{orange}{0.539}
& \color{black}{0.692}
\\
\cline{2- 9 }
\hline
\hline
\multirow{ 2 }{*}{{ Male }}
& { Male }
& \color{gray}{0.765}
& \color{gray}{0.775}
& \color{gray}{0.814}
& \color{gray}{0.716}
& \color{gray}{0.773}
& \color{gray}{ \textbf{0.823} }
& \color{gray}{0.764}
\\
\cline{2- 9 }
& { Female }
& \color{black}{0.781}
& \color{black}{0.788}
& \color{black}{ \textbf{0.823} }
& \color{black}{0.718}
& \color{black}{0.771}
& \color{black}{0.809}
& \color{black}{0.751}
\\
\cline{2- 9 }
\hline
\hline
\multirow{ 2 }{*}{{ Female }}
& { Female }
& \color{gray}{0.779}
& \color{gray}{0.788}
& \color{gray}{ \textbf{0.816} }
& \color{gray}{0.727}
& \color{gray}{0.765}
& \color{gray}{0.797}
& \color{gray}{0.766}
\\
\cline{2- 9 }
& { Male }
& \color{black}{0.762}
& \color{black}{0.763}
& \color{black}{0.804}
& \color{black}{0.726}
& \color{black}{0.754}
& \color{black}{ \textbf{0.812} }
& \color{black}{0.775}
\\
\cline{2- 9 }
\hline
\hline
\multirow{ 5 }{*}{{ Age 15-50yr }}
& { Age 15-50yr }
& \color{gray}{0.767}
& \color{gray}{0.783}
& \color{gray}{ \textbf{0.784} }
& \color{gray}{0.741}
& \color{gray}{0.767}
& \color{gray}{0.775}
& \color{gray}{0.717}
\\
\cline{2- 9 }
& { Age 50-60yr }
& \color{black}{0.757}
& \color{black}{0.761}
& \color{black}{0.790}
& \color{black}{0.747}
& \color{black}{0.748}
& \color{black}{ \textbf{0.793} }
& \color{black}{0.732}
\\
\cline{2- 9 }
& { Age 60-70yr }
& \color{black}{0.711}
& \color{black}{0.702}
& \color{black}{0.712}
& \color{black}{0.677}
& \color{black}{0.691}
& \color{black}{ \textbf{0.744} }
& \color{black}{0.668}
\\
\cline{2- 9 }
& { Age 70-80yr }
& \color{black}{0.661}
& \color{black}{0.640}
& \color{black}{ \textbf{0.706} }
& \color{black}{0.667}
& \color{black}{0.631}
& \color{black}{0.696}
& \color{black}{0.649}
\\
\cline{2- 9 }
& { Age 80+yr }
& \color{black}{0.611}
& \color{black}{0.591}
& \color{black}{0.623}
& \color{black}{0.609}
& \color{black}{0.614}
& \color{black}{ \textbf{0.667} }
& \color{black}{0.600}
\\
\cline{2- 9 }
\hline
\end{tabular}
    \caption{Performance of various algorithms as measured with AUC against the In-Hospital Mortality task. The IID numbers are in dark yellow. When the OOD performance drops to random guessing or worse, the value is colored Red. If the OOD performance happens to be significantly better than the IND performance, those values are colored Green. Within each row, the best performing algorithm's value is in bold.}
    \label{tab:mortauc}
\end{table}

\begin{table}[]
    \centering
    \small
        
\begin{tabular}{|l|l|c|c|c|c|c|c|c|}
\hline
\multicolumn{ 9 }{|c|}{ In Hospital Mortality  - ECE (smaller values are better)} \\
\hline
Train & Test & LogReg & GP & RF & MF & MLP & BRNN & SNGP\\
\hline
\multirow{ 3 }{*}{{ Adult }}
& { Adult }
& \color{black}{0.201}
& \color{black}{0.221}
& \color{black}{0.195}
& \color{black}{0.203}
& \color{black}{ \textbf{0.175} }
& \color{black}{0.196}
& \color{black}{0.216}
\\
\cline{2- 9 }
& { Neonate }
& \color{blue}{0.123}
& \color{blue}{0.129}
& \color{orange}{0.203}
& \color{orange}{0.236}
& \color{blue}{ \textbf{0.101} }
& \color{black}{0.204}
& \color{blue}{0.140}
\\
\cline{2- 9 }
& { Paediatric }
& \color{blue}{ \textbf{0.114} }
& \color{orange}{0.243}
& \color{black}{0.192}
& \color{blue}{0.152}
& \color{blue}{0.117}
& \color{blue}{0.177}
& \color{blue}{0.152}
\\
\cline{2- 9 }
\hline
\hline
\multirow{ 3 }{*}{{ Neonate }}
& { Neonate }
& \color{black}{0.010}
& \color{black}{0.013}
& \color{black}{ \textbf{0.008} }
& \color{black}{ \textbf{0.008} }
& \color{black}{0.030}
& \color{black}{0.014}
& \color{black}{0.327}
\\
\cline{2- 9 }
& { Adult }
& \color{orange}{0.370}
& \color{orange}{0.438}
& \color{orange}{0.239}
& \color{orange}{0.168}
& \color{orange}{ \textbf{0.127} }
& \color{orange}{0.129}
& \color{orange}{0.387}
\\
\cline{2- 9 }
& { Paediatric }
& \color{orange}{ \textbf{0.061} }
& \color{orange}{0.292}
& \color{orange}{0.112}
& \color{orange}{0.070}
& \color{orange}{ \textbf{0.061} }
& \color{orange}{0.064}
& \color{orange}{0.362}
\\
\cline{2- 9 }
\hline
\hline
\multirow{ 3 }{*}{{ Paediatric }}
& { Paediatric }
& \color{black}{0.102}
& \color{black}{0.113}
& \color{black}{ \textbf{0.098} }
& \color{black}{0.102}
& \color{black}{0.101}
& \color{black}{0.107}
& \color{black}{0.142}
\\
\cline{2- 9 }
& { Adult }
& \color{orange}{0.175}
& \color{orange}{0.232}
& \color{orange}{0.235}
& \color{orange}{0.170}
& \color{orange}{ \textbf{0.129} }
& \color{orange}{0.135}
& \color{orange}{0.273}
\\
\cline{2- 9 }
& { Neonate }
& \color{blue}{0.051}
& \color{blue}{0.037}
& \color{blue}{0.041}
& \color{blue}{0.043}
& \color{blue}{ \textbf{0.008} }
& \color{blue}{0.089}
& \color{blue}{0.091}
\\
\cline{2- 9 }
\hline
\hline
\multirow{ 2 }{*}{{ Male }}
& { Male }
& \color{black}{0.177}
& \color{black}{0.188}
& \color{black}{0.162}
& \color{black}{0.170}
& \color{black}{0.162}
& \color{black}{ \textbf{0.155} }
& \color{black}{0.166}
\\
\cline{2- 9 }
& { Female }
& \color{black}{0.178}
& \color{black}{0.186}
& \color{black}{0.160}
& \color{black}{0.169}
& \color{black}{0.162}
& \color{black}{ \textbf{0.157} }
& \color{black}{0.167}
\\
\cline{2- 9 }
\hline
\hline
\multirow{ 2 }{*}{{ Female }}
& { Female }
& \color{black}{0.179}
& \color{black}{0.184}
& \color{black}{ \textbf{0.162} }
& \color{black}{0.168}
& \color{black}{0.164}
& \color{black}{0.169}
& \color{black}{0.179}
\\
\cline{2- 9 }
& { Male }
& \color{black}{0.179}
& \color{black}{0.186}
& \color{black}{ \textbf{0.164} }
& \color{black}{0.170}
& \color{black}{ \textbf{0.164} }
& \color{black}{0.167}
& \color{black}{0.179}
\\
\cline{2- 9 }
\hline
\hline
\multirow{ 5 }{*}{{ Age 15-50yr }}
& { Age 15-50yr }
& \color{black}{0.122}
& \color{black}{0.148}
& \color{black}{0.119}
& \color{black}{0.119}
& \color{black}{0.139}
& \color{black}{0.129}
& \color{black}{ \textbf{0.115} }
\\
\cline{2- 9 }
& { Age 50-60yr }
& \color{orange}{0.143}
& \color{orange}{0.174}
& \color{orange}{0.185}
& \color{orange}{0.140}
& \color{orange}{0.162}
& \color{black}{ \textbf{0.137} }
& \color{orange}{0.141}
\\
\cline{2- 9 }
& { Age 60-70yr }
& \color{orange}{0.164}
& \color{orange}{0.201}
& \color{orange}{0.201}
& \color{orange}{ \textbf{0.156} }
& \color{orange}{0.186}
& \color{orange}{ \textbf{0.156} }
& \color{orange}{0.164}
\\
\cline{2- 9 }
& { Age 70-80yr }
& \color{orange}{0.190}
& \color{orange}{0.227}
& \color{orange}{0.220}
& \color{orange}{0.180}
& \color{orange}{0.212}
& \color{orange}{ \textbf{0.176} }
& \color{orange}{0.191}
\\
\cline{2- 9 }
& { Age 80+yr }
& \color{orange}{0.255}
& \color{orange}{0.315}
& \color{orange}{0.264}
& \color{orange}{0.228}
& \color{orange}{0.265}
& \color{orange}{ \textbf{0.225} }
& \color{orange}{0.290}
\\
\cline{2- 9 }
\hline
\end{tabular}
    \caption{Performance of various algorithms as measured with ECE against the In-Hospital Mortality task. The IID numbers are in Gray. When the OOD performance is significantly worse (or better), the value is colored Red (or Green). Within each row, the best performing algorithm's value is in bold.}
    \label{tab:mortece}
\end{table}

\begin{table}[]
    \centering
    \small

\begin{tabular}{|l|l|c|c|c|c|c|c|c|}
\hline
\multicolumn{ 9 }{|c|}{ In Hospital Mortality  - OOD} \\
\hline
Train & Test & LogReg & GP & RF & MF & MLP & BRNN & SNGP\\
\hline
\multirow{ 2 }{*}{{ Adult }}
& { Neonate }
& \color{black}{0.644}
& \color{black}{0.507}
& \color{black}{0.752}
& \color{black}{ \textbf{0.793} }
& \color{black}{0.678}
& \color{black}{0.694}
& \color{black}{0.620}
\\
\cline{2- 9 }
& { Paediatric }
& \color{black}{0.229}
& \color{black}{ \textbf{0.784} }
& \color{black}{0.684}
& \color{black}{0.544}
& \color{black}{0.456}
& \color{black}{0.536}
& \color{black}{0.309}
\\
\cline{2- 9 }
\hline
\hline
\multirow{ 2 }{*}{{ Neonate }}
& { Adult }
& \color{black}{0.576}
& \color{black}{0.999}
& \color{black}{0.914}
& \color{black}{0.976}
& \color{black}{0.000}
& \color{black}{0.020}
& \color{black}{ \textbf{1.000} }
\\
\cline{2- 9 }
& { Paediatric }
& \color{black}{0.162}
& \color{black}{0.966}
& \color{black}{0.853}
& \color{black}{0.950}
& \color{black}{0.047}
& \color{black}{0.327}
& \color{black}{ \textbf{0.989} }
\\
\cline{2- 9 }
\hline
\hline
\multirow{ 2 }{*}{{ Paediatric }}
& { Adult }
& \color{black}{0.242}
& \color{black}{0.746}
& \color{black}{ \textbf{0.799} }
& \color{black}{0.701}
& \color{black}{0.061}
& \color{black}{0.160}
& \color{black}{0.795}
\\
\cline{2- 9 }
& { Neonate }
& \color{black}{0.613}
& \color{black}{0.417}
& \color{black}{0.500}
& \color{black}{0.604}
& \color{black}{0.120}
& \color{black}{ \textbf{0.756} }
& \color{black}{0.536}
\\
\cline{2- 9 }
\hline
\hline
\multirow{ 1 }{*}{{ Male }}
& { Female }
& \color{black}{0.496}
& \color{black}{ \textbf{0.498} }
& \color{black}{0.492}
& \color{black}{0.493}
& \color{black}{0.494}
& \color{black}{0.494}
& \color{black}{0.496}
\\
\cline{2- 9 }
\hline
\hline
\multirow{ 1 }{*}{{ Female }}
& { Male }
& \color{black}{0.500}
& \color{black}{0.498}
& \color{black}{0.505}
& \color{black}{0.505}
& \color{black}{0.500}
& \color{black}{0.506}
& \color{black}{ \textbf{0.510} }
\\
\cline{2- 9 }
\hline
\hline
\multirow{ 4 }{*}{{ Age 15-50yr }}
& { Age 50-60yr }
& \color{black}{0.545}
& \color{black}{0.566}
& \color{black}{ \textbf{0.782} }
& \color{black}{0.511}
& \color{black}{0.535}
& \color{black}{0.490}
& \color{black}{0.558}
\\
\cline{2- 9 }
& { Age 60-70yr }
& \color{black}{0.563}
& \color{black}{0.619}
& \color{black}{ \textbf{0.774} }
& \color{black}{0.493}
& \color{black}{0.568}
& \color{black}{0.488}
& \color{black}{0.571}
\\
\cline{2- 9 }
& { Age 70-80yr }
& \color{black}{0.603}
& \color{black}{0.656}
& \color{black}{ \textbf{0.780} }
& \color{black}{0.518}
& \color{black}{0.587}
& \color{black}{0.496}
& \color{black}{0.618}
\\
\cline{2- 9 }
& { Age 80+yr }
& \color{black}{0.713}
& \color{black}{0.765}
& \color{black}{ \textbf{0.778} }
& \color{black}{0.537}
& \color{black}{0.656}
& \color{black}{0.488}
& \color{black}{0.726}
\\
\cline{2- 9 }
\hline
\end{tabular}
    \caption{OOD detection performance of various algorithms on the In-Hospital Mortality task. There are no rows corresponding to IND performance since OOD detection is not well defined in this case. Within each row, the best performing algorithm's value is in bold.}
    \label{tab:mortood}
\end{table}

The results of one of the tasks (In-Hospital Mortality) are presented in Tables \ref{tab:mortauc} (for AUC), \ref{tab:mortece} (for ECE) and \ref{tab:mortood} (for OOD). The metrics are described in Section~\ref{sec:metrics}. Each row corresponds to a specific combination of train and test set, and each column (starting from the third column) corresponds to a learning algorithm. In the AUC table (Table \ref{tab:mortauc}) and the ECE table (Table \ref{tab:mortece}), the IND rows have values colored in gray. The red colored values are those where an improvement in performance is desired, and the green colored values are those where the performance is, somewhat unexpectedly, better than expected. Within each row, the best performing algorithm's value is set in bold. We do not color code the OOD table (Table \ref{tab:mortood}) since interpreting those numbers is a little more nuanced and should only be viewed in conjunction with the corresponding values in the other two tables, due to reasons described in the following section. 

The remaining results for the other tasks are reported in the Appendix.

\section{Discussion}\label{sec:discussion}

In the results from our experiments, we broadly observe that among the algorithms we tested no algorithm dominates another in its performance across the board. We also observe that every model does particularly poorly in staying calibrated in OOD settings, with the exception of testing on Neonates.

Among the various algorithms tested in our experiments, a few of them are specifically designed with robustness to OOD inputs in mind. The SNGP, MF, and GP algorithms in particular are known to have stronger uncertainty estimation properties, even under distributional shift, relative to other algorithms. The SNGP algorithm in our experiment is essentially an MLP with additional GP layer as the final layer and spectral normalization in the fully connected layers. The SNGP algorithm is designed to be distance preserving at each of the individual layers in the deep model, with the hypothesis that it helps prevent collapsing of OOD and IND inputs at the final layer. What we observe is that, while in some cases these algorithms do perform better in OOD settings sometimes with respect to AUC score, the overall performance, especially with the ECE score has scope for improvement. Our hypothesis, and hope, is that as these algorithms get increasingly tested against EHR data modality, as they have been so far against images, improvements to the methods would result in increased robustness and better performance in such OOD settings.

It is also interesting to note that the difference between Male and Female distributions seem to not matter much for the downstream tasks, and the models are able to generalize over just fine with respect to both AUC and ECE. Indeed, this is also reflected in the fact that the OOD performance between these two subgroups is very close to 0.5. This example highlights the nuances in interpreting the OOD table, especially in isolation. While other works \citep{trustissues} highlight the fact that the models failing to distinguish Male vs Female distributions with appropriate levels predictive uncertainty and consider it as a failure mode, a more careful analysis shows that this might not be a failure in itself. When models are able to perfectly generalize from Male to Female and vice-versa with respect to both AUC and ECE metrics, the expectation for them to assign lower confidence (higher uncertainty) to the OOD examples is really undue, and even incorrect. Having a high OOD-AUC (i.e. assigning lower confidence to OOD examples) becomes a desiderata only when the model fails to generalize well to those examples. Our view is that while OOD detection performance can be interesting in some situations,  the AUC and ECE metrics carry most of the story in terms of the model's OOD behavior.

\begin{figure}
    \centering
    \includegraphics[scale=0.45]{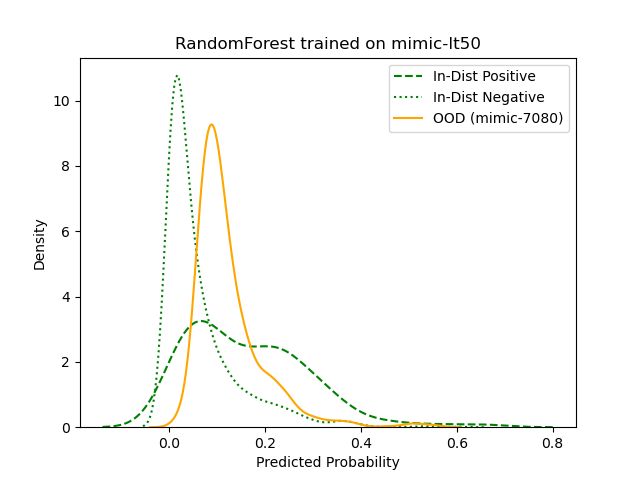}
    \includegraphics[scale=0.45]{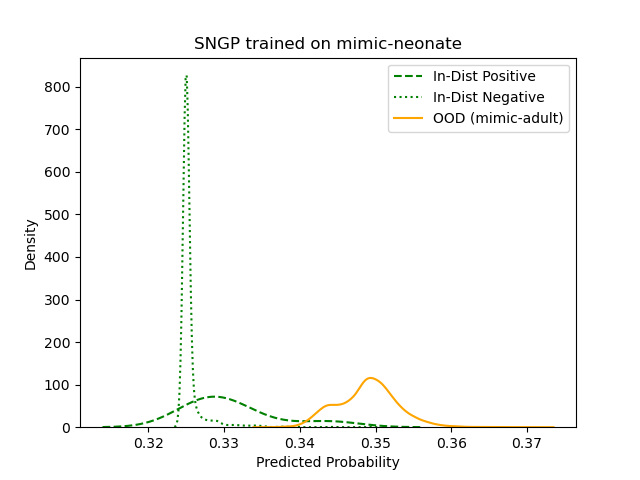}
    \caption{Two scenarios (models) plotting the In-Distribution postives, In-Distribution negatives, and OOD histograms (with smoothing) under class imbalance. The negative class is the majority and positives are few. In the left plot, the OOD class is ``in-between'' the positive and negative class. In the right plot, the OOD class is ``closer to 0.5'' than both the positives and negatives. Which of these two models is assigning ``lower confidence'' to the OOD examples?}
    \label{fig:confidence}
\end{figure}

A related consideration is about how to measure \emph{confidence} from a model's prediction that is in turn used to detect OOD examples. Among the class of models which involve a set of predictions for each input, such as ensembles \citep{deepensembles} or MC dropout \citep{mcdropout}, confidence is typically measured with the standard deviation (or mutual information \citep{trustissues}) using the set of predictions. A large standard deviation among the predictions represent higher uncertainty and vice versa.

For the second class of models, including many we have tested in this work, which output just one prediction per input, confidence is typically measured as the entropy of the predicted Bernoulli distribution (or Categorical for multi-class classification). A Bernoulli distribution with mean parameter $p=0.5$ has the highest entropy in its family, and therefore represents a prediction with least confidence, while those predictions with mean parameter close to 0 or 1 have lower entropy and hence represent predictions with high confidence. 

Among this second class of models, an alternate way of describing confidence would be to consider the model's discrimination ability, and inspect whether a given prediction is close to the threshold of maximum discrimination, or away from it (closer being ``less confident''). Here the threshold of maximum discrimination refers to the threshold value that maximizes the mean (either arithmetic or geometric) of the sensitivity and specificity of the model.

The question now is, is a low confidence prediction a Bernoulli distribution with high entropy, or is it a Bernoulli distribution whose mean is close to the threshold of maximum discrimination? This distinction is typically a moot point when there is a perfect class balance between the positive and negative classes (i.e. the marginal probability $p(y) = 0.5$). However when the classes are not well balanced, the two interpretations of confidence start to diverge, as illustrated in Figure \ref{fig:confidence}. The left plot is from a situation where the OOD examples are closer to the threshold of maximum discrimination. The right plot is from a different situation where the OOD examples have higher entropy overall (note the scale of the X-axes in both the plots differ), and thus achieves high OOD-AUC score. Yet, one might also reasonably interpret the right plot as the OOD examples being assigned a ``strongly positive'' score relative to the two IND classes, and hence the model is in fact being ``more confident'' on the OOD examples and therefore ought to have a  low OOD-AUC score under an appropriately chosen measure of confidence. These observations, in our opinion, makes the problem of OOD-detection itself a little less well defined under class imbalance situations for the class of models that use predictive entropy as a measure of confidence, in addition to it not being a very useful metric in isolation.

\section{Conclusion} \label{sec:conclusion}

In this work, we propose a benchmark, \beds, to evaluate an EHR ML model performance under distributional shift of the test data. We evaluate this benchmark on several algorithms, and find that no single algorithm demonstrates satisfactory robustness behavior over a wide range of OOD settings. We also find that no single algorithm works better than another across the board, including algorithms designed with OOD robustness in mind. While prior works have identified that most discriminative models are not reliable in detecting OOD examples reliably on medical tabular data, our work confirm this and in addition find that all the models we tested fare poorly in maintaining calibration under distributional shift of EHR data. This underscores the need for further research into robustness evaluation of EHR models, especially as these models are increasingly deployed in real-world clinical settings. 

\bibliography{main}

\clearpage
\newpage
\appendix 
\section*{Appendix}

\begin{table}[h]
    \centering
    \begin{tabular}{|c|c|c|}
    \hline
        Algorithm & Representation & Category \\
        \hline
        \hline
         Logistic Regression (LogReg)& Fixed-Length & Flat \\
         \hline
         Gaussian Process Classifier (GP) & Fixed-Length & Flat \\
         \hline
         \hline
         Random Forest (RF) & Fixed-Length & Tree-based \\
         \hline
         Mondrian Forest (MF) & Fixed-Length & Tree-based \\
         \hline
         \hline
         Multi Layer Perceptron (MLP) & Fixed-Length & Deep Learning \\
         \hline
         Bayesian RNN (BRNN)  & Sequential, Embeddings & Deep Learning \\
         \hline
         SNGP + MLP (SNGP) & Fixed-Length & Deep Learning \\
         \hline
    \end{tabular}
    \caption{Algorithms evaluated with \beds.}
    \label{tab:algos}
\end{table}

\begin{table}[h]
    \centering
    \resizebox{\textwidth}{!}{
    \begin{tabular}{|c|c|c|c|c|}
    \hline
    & MIMIC & PICDB & Harmonized To & Mapping Sources \\
    \hline
    Diagnostics & ICD-9 & ICD-10CN & ICD-9 & UMLS \\
    \hline
      Prescriptions   & National Drug Code & Text description & RXCUI & UMLS \\
         & (NDC) & & & MedEx \\
         \hline
         Lab Tests & LOINC+custom & LOINC+custom &  LOINC only & N/A \\
         \hline
         Vitals / Charts & Custom & Custom & MIMIC $\to$ PICDB & MIMIC-Extract \\
         \hline
         Input/Output & Custom & N/A & Volumes only & N/A \\
         Events & & & (ignore type) & \\
         \hline
    \end{tabular}
    }
    \caption{Details of harmonizing MIMIC-III and PICDB by category}
    \label{tab:harmonize}
\end{table}

\begin{table}[h]
    \centering
    \begin{tabular}{|l|r|r|r|r|}
    \hline
    & \cellalign{c|}{N} & \cellalign{c|}{Mort} & \cellalign{c|}{LoS3+} & \cellalign{c|}{LoS7+} \\
    \hline
    \hline
    MIMIC-adult & 36,909 & 13.3\% & 78.1\% & 42.9\% \\
    \hline
    PICDB-paed & 12,293 & 6.0\% & 91.2\% & 72.4\% \\
    \hline
    MIMIC-neonate & 7,651 & 0.5\% & 55.2\% & 29.8\% \\
    \hline
    \hline
    MIMIC-male & 25,004 & 10.5\% & 74.0\% & 40.6\% \\
    \hline
    MIMIC-female & 19,556 & 10.7\% & 74.2\% & 40.5\% \\
    \hline
    \hline
    MIMIC-lt50 & 7,795 & 7.2\% & 70.4\% & 38.5\% \\
    \hline
    MIMIC-5060 & 6,405 & 9.7\% & 77.6\% & 41.6\% \\
    \hline 
    MIMIC-6070 & 7,587 & 11.9\% & 81.9\% & 45.1\% \\
    \hline 
    MIMIC-7080 & 7,673 & 14.6\% & 82.3\% & 47.1\% \\
    \hline 
    MIMIC-gt80 & 7,449 & 19.9\% & 78.7\% & 42.4\% \\
    \hline
    \end{tabular}
    \caption{Sizes of data slices and their class balances for the three tasks}
    \label{tab:classbalance}
\end{table}

\begin{table}[h]
    \centering
    \begin{tabular}{|l|r|r|}
    \hline
       \cellalign{|c|}{Table}  & \cellalign{c}{Rows} & \cellalign{|c|}{Columns} \\
       \hline
        PATIENTS & 59,401 & 5 \\
        \hline
        ADMISSIONS & 72,425 & 6 \\
        \hline
        CHARTEVENTS & 45,577,975 & 7 \\
        \hline
        INPUTEVENTS & 16,068,433 & 5 \\
        \hline
        OUTPUTEVENTS & 4,300,561& 5 \\
        \hline
        LABEVENTS & 34,306,923 & 7 \\
        \hline
        PRESCRIPTIONS & 5,320,452 & 7 \\
        \hline
        DIAGNOSES\_ICD & 664,116 & 4 \\
        \hline
    \end{tabular}
    \caption{Summary of the resulting dataset after harmonizing and pre-processing the combined MIMIC-III and PICDB datasets.}
    \label{tab:resultingdata}
\end{table}

\begin{table}[h]
    \centering
       % \resizebox{\textwidth}{!}{

\begin{tabular}{|l|l|c|c|c|c|c|c|c|}
\hline
\multicolumn{ 9 }{|c|}{ Length of Stay 3+ days  - AUC} \\
\hline
Train & Test & LogReg & GP & RF & MF & MLP & BRNN & SNGP\\
\hline
\multirow{ 3 }{*}{{ Adult }}
& { Adult }
& \color{gray}{0.681}
& \color{gray}{0.685}
& \color{gray}{0.743}
& \color{gray}{0.626}
& \color{gray}{0.684}
& \color{gray}{ \textbf{0.756} }
& \color{gray}{0.679}
\\
\cline{2- 9 }
& { Neonate }
& \color{orange}{0.511}
& \color{orange}{0.305}
& \color{orange}{0.233}
& \color{orange}{0.332}
& \color{orange}{0.214}
& \color{orange}{0.196}
& \color{black}{ \textbf{0.583} }
\\
\cline{2- 9 }
& { Paediatric }
& \color{orange}{0.366}
& \color{black}{ \textbf{0.649} }
& \color{orange}{0.514}
& \color{orange}{0.478}
& \color{orange}{0.364}
& \color{orange}{0.398}
& \color{black}{0.582}
\\
\cline{2- 9 }
\hline
\hline
\multirow{ 3 }{*}{{ Neonate }}
& { Neonate }
& \color{gray}{0.888}
& \color{gray}{0.883}
& \color{gray}{0.876}
& \color{gray}{0.837}
& \color{gray}{0.884}
& \color{gray}{ \textbf{0.910} }
& \color{gray}{0.887}
\\
\cline{2- 9 }
& { Adult }
& \color{black}{0.551}
& \color{black}{0.576}
& \color{orange}{0.365}
& \color{black}{ \textbf{0.699} }
& \color{black}{0.526}
& \color{orange}{0.476}
& \color{orange}{0.435}
\\
\cline{2- 9 }
& { Paediatric }
& \color{orange}{0.375}
& \color{black}{ \textbf{0.588} }
& \color{orange}{0.393}
& \color{orange}{0.449}
& \color{orange}{0.306}
& \color{orange}{0.440}
& \color{black}{0.569}
\\
\cline{2- 9 }
\hline
\hline
\multirow{ 3 }{*}{{ Paediatric }}
& { Paediatric }
& \color{gray}{0.766}
& \color{gray}{0.790}
& \color{gray}{ \textbf{0.801} }
& \color{gray}{0.740}
& \color{gray}{0.789}
& \color{gray}{0.767}
& \color{gray}{0.767}
\\
\cline{2- 9 }
& { Adult }
& \color{orange}{0.401}
& \color{orange}{0.410}
& \color{orange}{0.377}
& \color{black}{ \textbf{0.584} }
& \color{orange}{0.444}
& \color{orange}{0.495}
& \color{orange}{0.419}
\\
\cline{2- 9 }
& { Neonate }
& \color{black}{0.680}
& \color{orange}{0.522}
& \color{orange}{0.463}
& \color{orange}{0.372}
& \color{black}{0.622}
& \color{blue}{ \textbf{0.847} }
& \color{black}{0.711}
\\
\cline{2- 9 }
\hline
\hline
\multirow{ 2 }{*}{{ Male }}
& { Male }
& \color{gray}{0.718}
& \color{gray}{0.760}
& \color{gray}{0.795}
& \color{gray}{0.709}
& \color{gray}{0.747}
& \color{gray}{ \textbf{0.800} }
& \color{gray}{0.739}
\\
\cline{2- 9 }
& { Female }
& \color{black}{0.729}
& \color{black}{0.768}
& \color{black}{ \textbf{0.803} }
& \color{black}{0.692}
& \color{black}{0.756}
& \color{black}{0.791}
& \color{black}{0.750}
\\
\cline{2- 9 }
\hline
\hline
\multirow{ 2 }{*}{{ Female }}
& { Female }
& \color{gray}{0.733}
& \color{gray}{0.772}
& \color{gray}{ \textbf{0.803} }
& \color{gray}{0.713}
& \color{gray}{0.741}
& \color{gray}{0.780}
& \color{gray}{0.750}
\\
\cline{2- 9 }
& { Male }
& \color{black}{0.725}
& \color{black}{0.772}
& \color{black}{ \textbf{0.798} }
& \color{black}{0.719}
& \color{black}{0.748}
& \color{black}{0.792}
& \color{black}{0.754}
\\
\cline{2- 9 }
\hline
\hline
\multirow{ 5 }{*}{{ Age 15-50yr }}
& { Age 15-50yr }
& \color{gray}{0.722}
& \color{gray}{0.731}
& \color{gray}{ \textbf{0.757} }
& \color{gray}{0.669}
& \color{gray}{0.702}
& \color{gray}{0.753}
& \color{gray}{0.726}
\\
\cline{2- 9 }
& { Age 50-60yr }
& \color{black}{0.699}
& \color{black}{0.701}
& \color{black}{ \textbf{0.772} }
& \color{black}{0.681}
& \color{black}{0.718}
& \color{black}{0.758}
& \color{black}{0.710}
\\
\cline{2- 9 }
& { Age 60-70yr }
& \color{black}{0.670}
& \color{black}{0.698}
& \color{black}{ \textbf{0.752} }
& \color{black}{0.666}
& \color{black}{0.670}
& \color{black}{0.728}
& \color{black}{0.681}
\\
\cline{2- 9 }
& { Age 70-80yr }
& \color{black}{0.640}
& \color{black}{0.654}
& \color{black}{ \textbf{0.716} }
& \color{black}{0.630}
& \color{black}{0.641}
& \color{black}{0.680}
& \color{black}{0.656}
\\
\cline{2- 9 }
& { Age 80+yr }
& \color{black}{0.558}
& \color{black}{0.575}
& \color{black}{ \textbf{0.713} }
& \color{black}{0.624}
& \color{black}{0.623}
& \color{black}{0.672}
& \color{black}{0.589}
\\
\cline{2- 9 }
\hline
\end{tabular}

%}
    %\caption{}
    %\label{}
\end{table}

\begin{table}[h]
    \centering
   % \resizebox{\textwidth}{!}{

\begin{tabular}{|l|l|c|c|c|c|c|c|c|}
\hline
\multicolumn{ 9 }{|c|}{ Length of Stay 3+ days  - ECE} \\
\hline
Train & Test & LogReg & GP & RF & MF & MLP & BRNN & SNGP\\
\hline
\multirow{ 3 }{*}{{ Adult }}
& { Adult }
& \color{black}{0.330}
& \color{black}{0.331}
& \color{black}{0.290}
& \color{black}{0.318}
& \color{black}{0.323}
& \color{black}{ \textbf{0.285} }
& \color{black}{0.312}
\\
\cline{2- 9 }
& { Neonate }
& \color{orange}{0.468}
& \color{orange}{0.476}
& \color{orange}{0.521}
& \color{orange}{0.477}
& \color{orange}{0.484}
& \color{orange}{0.584}
& \color{orange}{ \textbf{0.459} }
\\
\cline{2- 9 }
& { Paediatric }
& \color{blue}{0.230}
& \color{blue}{0.302}
& \color{blue}{0.263}
& \color{blue}{0.221}
& \color{orange}{0.367}
& \color{orange}{0.448}
& \color{blue}{ \textbf{0.202} }
\\
\cline{2- 9 }
\hline
\hline
\multirow{ 3 }{*}{{ Neonate }}
& { Neonate }
& \color{black}{0.318}
& \color{black}{0.284}
& \color{black}{0.262}
& \color{black}{0.266}
& \color{black}{0.263}
& \color{black}{ \textbf{0.250} }
& \color{black}{0.289}
\\
\cline{2- 9 }
& { Adult }
& \color{orange}{0.578}
& \color{orange}{0.456}
& \color{orange}{0.386}
& \color{orange}{0.495}
& \color{blue}{ \textbf{0.213} }
& \color{orange}{0.784}
& \color{orange}{0.322}
\\
\cline{2- 9 }
& { Paediatric }
& \color{orange}{0.415}
& \color{orange}{0.399}
& \color{orange}{0.515}
& \color{orange}{0.367}
& \color{black}{0.267}
& \color{orange}{0.511}
& \color{blue}{ \textbf{0.220} }
\\
\cline{2- 9 }
\hline
\hline
\multirow{ 3 }{*}{{ Paediatric }}
& { Paediatric }
& \color{black}{0.150}
& \color{black}{0.158}
& \color{black}{0.148}
& \color{black}{ \textbf{0.144} }
& \color{black}{0.179}
& \color{black}{0.149}
& \color{black}{0.160}
\\
\cline{2- 9 }
& { Adult }
& \color{orange}{0.272}
& \color{orange}{0.332}
& \color{orange}{0.331}
& \color{orange}{0.272}
& \color{orange}{0.244}
& \color{orange}{ \textbf{0.217} }
& \color{orange}{0.275}
\\
\cline{2- 9 }
& { Neonate }
& \color{orange}{0.447}
& \color{orange}{0.448}
& \color{orange}{0.452}
& \color{orange}{0.464}
& \color{orange}{0.446}
& \color{orange}{ \textbf{0.417} }
& \color{orange}{0.447}
\\
\cline{2- 9 }
\hline
\hline
\multirow{ 2 }{*}{{ Male }}
& { Male }
& \color{black}{0.332}
& \color{black}{0.336}
& \color{black}{0.290}
& \color{black}{0.318}
& \color{black}{0.343}
& \color{black}{ \textbf{0.283} }
& \color{black}{0.319}
\\
\cline{2- 9 }
& { Female }
& \color{black}{0.329}
& \color{black}{0.335}
& \color{black}{0.288}
& \color{black}{0.319}
& \color{black}{0.340}
& \color{black}{ \textbf{0.284} }
& \color{black}{0.315}
\\
\cline{2- 9 }
\hline
\hline
\multirow{ 2 }{*}{{ Female }}
& { Female }
& \color{black}{0.330}
& \color{black}{0.336}
& \color{black}{ \textbf{0.290} }
& \color{black}{0.318}
& \color{black}{0.330}
& \color{black}{0.300}
& \color{black}{0.324}
\\
\cline{2- 9 }
& { Male }
& \color{black}{0.332}
& \color{black}{0.337}
& \color{black}{ \textbf{0.292} }
& \color{black}{0.316}
& \color{black}{0.330}
& \color{black}{0.297}
& \color{black}{0.323}
\\
\cline{2- 9 }
\hline
\hline
\multirow{ 5 }{*}{{ Age 15-50yr }}
& { Age 15-50yr }
& \color{black}{0.374}
& \color{black}{0.382}
& \color{black}{0.348}
& \color{black}{0.391}
& \color{black}{0.379}
& \color{black}{ \textbf{0.346} }
& \color{black}{0.358}
\\
\cline{2- 9 }
& { Age 50-60yr }
& \color{blue}{0.335}
& \color{blue}{0.351}
& \color{blue}{0.330}
& \color{blue}{0.354}
& \color{blue}{0.342}
& \color{blue}{ \textbf{0.293} }
& \color{blue}{0.311}
\\
\cline{2- 9 }
& { Age 60-70yr }
& \color{blue}{0.310}
& \color{blue}{0.323}
& \color{blue}{0.312}
& \color{blue}{0.341}
& \color{blue}{0.328}
& \color{blue}{ \textbf{0.260} }
& \color{blue}{0.286}
\\
\cline{2- 9 }
& { Age 70-80yr }
& \color{blue}{0.302}
& \color{blue}{0.313}
& \color{blue}{0.316}
& \color{blue}{0.340}
& \color{blue}{0.327}
& \color{blue}{ \textbf{0.244} }
& \color{blue}{0.273}
\\
\cline{2- 9 }
& { Age 80+yr }
& \color{blue}{0.300}
& \color{blue}{0.319}
& \color{black}{0.348}
& \color{blue}{0.368}
& \color{blue}{0.347}
& \color{blue}{ \textbf{0.264} }
& \color{blue}{0.298}
\\
\cline{2- 9 }
\hline
\end{tabular}
%}
\end{table}

\begin{table}[h]
    \centering
      %  \resizebox{\textwidth}{!}{

\begin{tabular}{|l|l|c|c|c|c|c|c|c|}
\hline
\multicolumn{ 9 }{|c|}{ Length of Stay 3+ days  - OOD} \\
\hline
Train & Test & LogReg & GP & RF & MF & MLP & BRNN & SNGP\\
\hline
\multirow{ 2 }{*}{{ Adult }}
& { Neonate }
& \color{black}{0.560}
& \color{black}{0.313}
& \color{black}{0.542}
& \color{black}{0.246}
& \color{black}{0.310}
& \color{black}{ \textbf{0.827} }
& \color{black}{0.511}
\\
\cline{2- 9 }
& { Paediatric }
& \color{black}{0.185}
& \color{black}{0.600}
& \color{black}{0.542}
& \color{black}{0.345}
& \color{black}{0.719}
& \color{black}{ \textbf{0.833} }
& \color{black}{0.221}
\\
\cline{2- 9 }
\hline
\hline
\multirow{ 2 }{*}{{ Neonate }}
& { Adult }
& \color{black}{0.268}
& \color{black}{ \textbf{0.926} }
& \color{black}{0.728}
& \color{black}{0.901}
& \color{black}{0.002}
& \color{black}{0.015}
& \color{black}{0.356}
\\
\cline{2- 9 }
& { Paediatric }
& \color{black}{0.530}
& \color{black}{0.839}
& \color{black}{ \textbf{0.883} }
& \color{black}{0.837}
& \color{black}{0.401}
& \color{black}{0.800}
& \color{black}{0.275}
\\
\cline{2- 9 }
\hline
\hline
\multirow{ 2 }{*}{{ Paediatric }}
& { Adult }
& \color{black}{0.366}
& \color{black}{ \textbf{0.828} }
& \color{black}{0.788}
& \color{black}{0.704}
& \color{black}{0.190}
& \color{black}{0.066}
& \color{black}{0.371}
\\
\cline{2- 9 }
& { Neonate }
& \color{black}{0.444}
& \color{black}{0.272}
& \color{black}{0.441}
& \color{black}{ \textbf{0.539} }
& \color{black}{0.177}
& \color{black}{0.298}
& \color{black}{0.441}
\\
\cline{2- 9 }
\hline
\hline
\multirow{ 1 }{*}{{ Male }}
& { Female }
& \color{black}{ \textbf{0.501} }
& \color{black}{0.500}
& \color{black}{0.497}
& \color{black}{0.491}
& \color{black}{0.499}
& \color{black}{ \textbf{0.501} }
& \color{black}{0.495}
\\
\cline{2- 9 }
\hline
\hline
\multirow{ 1 }{*}{{ Female }}
& { Male }
& \color{black}{0.495}
& \color{black}{0.496}
& \color{black}{ \textbf{0.502} }
& \color{black}{0.499}
& \color{black}{0.499}
& \color{black}{ \textbf{0.502} }
& \color{black}{0.500}
\\
\cline{2- 9 }
\hline
\hline
\multirow{ 4 }{*}{{ Age 15-50yr }}
& { Age 50-60yr }
& \color{black}{0.422}
& \color{black}{0.434}
& \color{black}{ \textbf{0.514} }
& \color{black}{0.475}
& \color{black}{0.469}
& \color{black}{0.395}
& \color{black}{0.458}
\\
\cline{2- 9 }
& { Age 60-70yr }
& \color{black}{0.361}
& \color{black}{0.366}
& \color{black}{ \textbf{0.477} }
& \color{black}{0.451}
& \color{black}{0.435}
& \color{black}{0.313}
& \color{black}{0.419}
\\
\cline{2- 9 }
& { Age 70-80yr }
& \color{black}{0.345}
& \color{black}{0.345}
& \color{black}{ \textbf{0.490} }
& \color{black}{0.458}
& \color{black}{0.444}
& \color{black}{0.261}
& \color{black}{0.430}
\\
\cline{2- 9 }
& { Age 80+yr }
& \color{black}{0.255}
& \color{black}{0.278}
& \color{black}{ \textbf{0.546} }
& \color{black}{0.504}
& \color{black}{0.437}
& \color{black}{0.237}
& \color{black}{0.392}
\\
\cline{2- 9 }
\hline
\end{tabular}

%}
\end{table}

\begin{table}[h]
    \centering
       % \resizebox{\textwidth}{!}{

\begin{tabular}{|l|l|c|c|c|c|c|c|c|}
\hline
\multicolumn{ 9 }{|c|}{ Length of Stay 7+ days  - AUC} \\
\hline
Train & Test & LogReg & GP & RF & MF & MLP & BRNN & SNGP\\
\hline
\multirow{ 3 }{*}{{ Adult }}
& { Adult }
& \color{gray}{0.653}
& \color{gray}{0.658}
& \color{gray}{ \textbf{0.723} }
& \color{gray}{0.620}
& \color{gray}{0.647}
& \color{gray}{0.715}
& \color{gray}{0.639}
\\
\cline{2- 9 }
& { Neonate }
& \color{black}{0.554}
& \color{orange}{0.487}
& \color{black}{ \textbf{0.664} }
& \color{black}{0.655}
& \color{orange}{0.326}
& \color{orange}{0.205}
& \color{orange}{0.381}
\\
\cline{2- 9 }
& { Paediatric }
& \color{black}{0.539}
& \color{orange}{0.461}
& \color{orange}{0.444}
& \color{black}{ \textbf{0.583} }
& \color{black}{0.535}
& \color{black}{0.560}
& \color{orange}{0.457}
\\
\cline{2- 9 }
\hline
\hline
\multirow{ 3 }{*}{{ Neonate }}
& { Neonate }
& \color{gray}{0.913}
& \color{gray}{0.915}
& \color{gray}{0.898}
& \color{gray}{0.873}
& \color{gray}{0.916}
& \color{gray}{ \textbf{0.920} }
& \color{gray}{0.912}
\\
\cline{2- 9 }
& { Adult }
& \color{orange}{0.485}
& \color{black}{0.570}
& \color{orange}{0.502}
& \color{black}{0.567}
& \color{black}{ \textbf{0.586} }
& \color{orange}{0.482}
& \color{black}{0.538}
\\
\cline{2- 9 }
& { Paediatric }
& \color{orange}{0.507}
& \color{orange}{0.493}
& \color{orange}{0.401}
& \color{orange}{0.445}
& \color{orange}{0.451}
& \color{black}{ \textbf{0.529} }
& \color{orange}{0.503}
\\
\cline{2- 9 }
\hline
\hline
\multirow{ 3 }{*}{{ Paediatric }}
& { Paediatric }
& \color{gray}{0.721}
& \color{gray}{ \textbf{0.745} }
& \color{gray}{0.731}
& \color{gray}{0.695}
& \color{gray}{0.713}
& \color{gray}{0.719}
& \color{gray}{0.717}
\\
\cline{2- 9 }
& { Adult }
& \color{orange}{0.445}
& \color{orange}{0.413}
& \color{orange}{0.504}
& \color{black}{ \textbf{0.563} }
& \color{orange}{0.434}
& \color{orange}{0.502}
& \color{orange}{0.442}
\\
\cline{2- 9 }
& { Neonate }
& \color{blue}{0.815}
& \color{black}{0.644}
& \color{orange}{0.485}
& \color{orange}{0.292}
& \color{orange}{0.468}
& \color{black}{0.766}
& \color{blue}{ \textbf{0.825} }
\\
\cline{2- 9 }
\hline
\hline
\multirow{ 2 }{*}{{ Male }}
& { Male }
& \color{gray}{0.698}
& \color{gray}{0.732}
& \color{gray}{ \textbf{0.763} }
& \color{gray}{0.674}
& \color{gray}{0.714}
& \color{gray}{0.749}
& \color{gray}{0.699}
\\
\cline{2- 9 }
& { Female }
& \color{black}{0.698}
& \color{black}{0.733}
& \color{black}{ \textbf{0.763} }
& \color{black}{0.679}
& \color{black}{0.720}
& \color{black}{0.757}
& \color{black}{0.698}
\\
\cline{2- 9 }
\hline
\hline
\multirow{ 2 }{*}{{ Female }}
& { Female }
& \color{gray}{0.697}
& \color{gray}{0.731}
& \color{gray}{ \textbf{0.762} }
& \color{gray}{0.691}
& \color{gray}{0.717}
& \color{gray}{0.742}
& \color{gray}{0.707}
\\
\cline{2- 9 }
& { Male }
& \color{black}{0.698}
& \color{black}{0.733}
& \color{black}{ \textbf{0.765} }
& \color{black}{0.690}
& \color{black}{0.719}
& \color{black}{0.738}
& \color{black}{0.703}
\\
\cline{2- 9 }
\hline
\hline
\multirow{ 5 }{*}{{ Age 15-50yr }}
& { Age 15-50yr }
& \color{gray}{0.701}
& \color{gray}{0.700}
& \color{gray}{ \textbf{0.753} }
& \color{gray}{0.665}
& \color{gray}{0.708}
& \color{gray}{0.731}
& \color{gray}{0.709}
\\
\cline{2- 9 }
& { Age 50-60yr }
& \color{black}{0.678}
& \color{black}{0.683}
& \color{black}{ \textbf{0.712} }
& \color{black}{0.637}
& \color{black}{0.662}
& \color{black}{0.704}
& \color{black}{0.674}
\\
\cline{2- 9 }
& { Age 60-70yr }
& \color{black}{0.650}
& \color{black}{0.672}
& \color{black}{ \textbf{0.743} }
& \color{black}{0.634}
& \color{black}{0.665}
& \color{black}{0.690}
& \color{black}{0.678}
\\
\cline{2- 9 }
& { Age 70-80yr }
& \color{black}{0.621}
& \color{black}{0.651}
& \color{black}{ \textbf{0.694} }
& \color{black}{0.628}
& \color{black}{0.614}
& \color{black}{0.677}
& \color{black}{0.644}
\\
\cline{2- 9 }
& { Age 80+yr }
& \color{black}{0.550}
& \color{black}{0.542}
& \color{black}{ \textbf{0.688} }
& \color{black}{0.583}
& \color{black}{0.572}
& \color{black}{0.651}
& \color{black}{0.556}
\\
\cline{2- 9 }
\hline
\end{tabular}

%}
\end{table}

\begin{table}[h]
    \centering
     %   \resizebox{\textwidth}{!}{

\begin{tabular}{|l|l|c|c|c|c|c|c|c|}
\hline
\multicolumn{ 9 }{|c|}{ Length of Stay 7+ days  - ECE} \\
\hline
Train & Test & LogReg & GP & RF & MF & MLP & BRNN & SNGP\\
\hline
\multirow{ 3 }{*}{{ Adult }}
& { Adult }
& \color{black}{0.474}
& \color{black}{0.458}
& \color{black}{ \textbf{0.420} }
& \color{black}{0.459}
& \color{black}{0.455}
& \color{black}{ \textbf{0.420} }
& \color{black}{0.479}
\\
\cline{2- 9 }
& { Neonate }
& \color{black}{0.473}
& \color{orange}{0.502}
& \color{orange}{0.468}
& \color{black}{ \textbf{0.455} }
& \color{orange}{0.533}
& \color{orange}{0.517}
& \color{orange}{0.506}
\\
\cline{2- 9 }
& { Paediatric }
& \color{orange}{0.545}
& \color{orange}{ \textbf{0.511} }
& \color{orange}{0.514}
& \color{orange}{0.530}
& \color{orange}{0.534}
& \color{orange}{0.559}
& \color{orange}{0.515}
\\
\cline{2- 9 }
\hline
\hline
\multirow{ 3 }{*}{{ Neonate }}
& { Neonate }
& \color{black}{0.250}
& \color{black}{0.215}
& \color{black}{0.210}
& \color{black}{0.212}
& \color{black}{ \textbf{0.197} }
& \color{black}{0.201}
& \color{black}{0.212}
\\
\cline{2- 9 }
& { Adult }
& \color{orange}{0.447}
& \color{orange}{0.492}
& \color{orange}{0.497}
& \color{orange}{0.465}
& \color{orange}{ \textbf{0.427} }
& \color{orange}{0.442}
& \color{orange}{0.511}
\\
\cline{2- 9 }
& { Paediatric }
& \color{orange}{0.609}
& \color{orange}{0.517}
& \color{orange}{0.569}
& \color{orange}{0.553}
& \color{orange}{0.463}
& \color{orange}{0.640}
& \color{orange}{ \textbf{0.415} }
\\
\cline{2- 9 }
\hline
\hline
\multirow{ 3 }{*}{{ Paediatric }}
& { Paediatric }
& \color{black}{0.368}
& \color{black}{0.356}
& \color{black}{ \textbf{0.348} }
& \color{black}{ \textbf{0.348} }
& \color{black}{0.354}
& \color{black}{0.361}
& \color{black}{0.357}
\\
\cline{2- 9 }
& { Adult }
& \color{orange}{0.541}
& \color{orange}{ \textbf{0.528} }
& \color{orange}{0.532}
& \color{orange}{0.529}
& \color{orange}{0.569}
& \color{orange}{0.555}
& \color{orange}{0.541}
\\
\cline{2- 9 }
& { Neonate }
& \color{orange}{ \textbf{0.604} }
& \color{orange}{0.651}
& \color{orange}{0.624}
& \color{orange}{0.656}
& \color{orange}{0.636}
& \color{orange}{0.609}
& \color{orange}{0.621}
\\
\cline{2- 9 }
\hline
\hline
\multirow{ 2 }{*}{{ Male }}
& { Male }
& \color{black}{0.451}
& \color{black}{0.424}
& \color{black}{ \textbf{0.380} }
& \color{black}{0.419}
& \color{black}{0.413}
& \color{black}{0.388}
& \color{black}{0.430}
\\
\cline{2- 9 }
& { Female }
& \color{black}{0.449}
& \color{black}{0.421}
& \color{black}{ \textbf{0.379} }
& \color{black}{0.415}
& \color{black}{0.409}
& \color{black}{0.384}
& \color{black}{0.429}
\\
\cline{2- 9 }
\hline
\hline
\multirow{ 2 }{*}{{ Female }}
& { Female }
& \color{black}{0.449}
& \color{black}{0.421}
& \color{black}{ \textbf{0.378} }
& \color{black}{0.409}
& \color{black}{0.407}
& \color{black}{0.385}
& \color{black}{0.428}
\\
\cline{2- 9 }
& { Male }
& \color{black}{0.451}
& \color{black}{0.423}
& \color{black}{ \textbf{0.380} }
& \color{black}{0.413}
& \color{black}{0.410}
& \color{black}{0.390}
& \color{black}{0.428}
\\
\cline{2- 9 }
\hline
\hline
\multirow{ 5 }{*}{{ Age 15-50yr }}
& { Age 15-50yr }
& \color{black}{0.451}
& \color{black}{0.427}
& \color{black}{ \textbf{0.388} }
& \color{black}{0.431}
& \color{black}{0.422}
& \color{black}{0.411}
& \color{black}{0.429}
\\
\cline{2- 9 }
& { Age 50-60yr }
& \color{orange}{0.462}
& \color{black}{0.440}
& \color{orange}{ \textbf{0.422} }
& \color{orange}{0.448}
& \color{orange}{0.439}
& \color{black}{0.428}
& \color{orange}{0.447}
\\
\cline{2- 9 }
& { Age 60-70yr }
& \color{orange}{0.466}
& \color{orange}{0.449}
& \color{orange}{ \textbf{0.412} }
& \color{orange}{0.449}
& \color{orange}{0.439}
& \color{orange}{0.435}
& \color{orange}{0.448}
\\
\cline{2- 9 }
& { Age 70-80yr }
& \color{orange}{0.479}
& \color{orange}{0.461}
& \color{orange}{ \textbf{0.436} }
& \color{orange}{0.461}
& \color{orange}{0.463}
& \color{orange}{0.446}
& \color{orange}{0.461}
\\
\cline{2- 9 }
& { Age 80+yr }
& \color{orange}{0.478}
& \color{orange}{0.491}
& \color{orange}{ \textbf{0.431} }
& \color{orange}{0.462}
& \color{orange}{0.466}
& \color{orange}{0.459}
& \color{orange}{0.476}
\\
\cline{2- 9 }
\hline
\end{tabular}

%}
\end{table}

\begin{table}[h]
    \centering
     %   \resizebox{\textwidth}{!}{

\begin{tabular}{|l|l|c|c|c|c|c|c|c|}
\hline
\multicolumn{ 9 }{|c|}{ Length of Stay 7+ days  - OOD} \\
\hline
Train & Test & LogReg & GP & RF & MF & MLP & BRNN & SNGP\\
\hline
\multirow{ 2 }{*}{{ Adult }}
& { Neonate }
& \color{black}{0.655}
& \color{black}{0.814}
& \color{black}{0.766}
& \color{black}{0.575}
& \color{black}{0.637}
& \color{black}{0.439}
& \color{black}{ \textbf{0.850} }
\\
\cline{2- 9 }
& { Paediatric }
& \color{black}{0.350}
& \color{black}{ \textbf{0.860} }
& \color{black}{0.785}
& \color{black}{0.615}
& \color{black}{0.562}
& \color{black}{0.542}
& \color{black}{0.714}
\\
\cline{2- 9 }
\hline
\hline
\multirow{ 2 }{*}{{ Neonate }}
& { Adult }
& \color{black}{0.054}
& \color{black}{ \textbf{0.957} }
& \color{black}{0.950}
& \color{black}{0.865}
& \color{black}{0.157}
& \color{black}{0.151}
& \color{black}{0.821}
\\
\cline{2- 9 }
& { Paediatric }
& \color{black}{0.607}
& \color{black}{ \textbf{0.966} }
& \color{black}{0.876}
& \color{black}{0.906}
& \color{black}{0.280}
& \color{black}{0.608}
& \color{black}{0.630}
\\
\cline{2- 9 }
\hline
\hline
\multirow{ 2 }{*}{{ Paediatric }}
& { Adult }
& \color{black}{0.476}
& \color{black}{ \textbf{0.743} }
& \color{black}{0.476}
& \color{black}{0.524}
& \color{black}{0.372}
& \color{black}{0.053}
& \color{black}{0.547}
\\
\cline{2- 9 }
& { Neonate }
& \color{black}{0.243}
& \color{black}{0.080}
& \color{black}{ \textbf{0.391} }
& \color{black}{0.309}
& \color{black}{0.275}
& \color{black}{0.214}
& \color{black}{0.303}
\\
\cline{2- 9 }
\hline
\hline
\multirow{ 1 }{*}{{ Male }}
& { Female }
& \color{black}{0.492}
& \color{black}{0.496}
& \color{black}{0.496}
& \color{black}{0.494}
& \color{black}{0.500}
& \color{black}{ \textbf{0.502} }
& \color{black}{0.497}
\\
\cline{2- 9 }
\hline
\hline
\multirow{ 1 }{*}{{ Female }}
& { Male }
& \color{black}{0.499}
& \color{black}{0.507}
& \color{black}{0.504}
& \color{black}{ \textbf{0.512} }
& \color{black}{0.505}
& \color{black}{0.504}
& \color{black}{0.496}
\\
\cline{2- 9 }
\hline
\hline
\multirow{ 4 }{*}{{ Age 15-50yr }}
& { Age 50-60yr }
& \color{black}{0.501}
& \color{black}{0.546}
& \color{black}{ \textbf{0.574} }
& \color{black}{0.489}
& \color{black}{0.499}
& \color{black}{0.531}
& \color{black}{0.531}
\\
\cline{2- 9 }
& { Age 60-70yr }
& \color{black}{0.507}
& \color{black}{ \textbf{0.571} }
& \color{black}{0.567}
& \color{black}{0.496}
& \color{black}{0.493}
& \color{black}{0.533}
& \color{black}{0.549}
\\
\cline{2- 9 }
& { Age 70-80yr }
& \color{black}{0.516}
& \color{black}{ \textbf{0.607} }
& \color{black}{0.591}
& \color{black}{0.503}
& \color{black}{0.520}
& \color{black}{0.581}
& \color{black}{0.562}
\\
\cline{2- 9 }
& { Age 80+yr }
& \color{black}{0.603}
& \color{black}{0.565}
& \color{black}{0.583}
& \color{black}{0.511}
& \color{black}{0.547}
& \color{black}{0.600}
& \color{black}{ \textbf{0.616} }
\\
\cline{2- 9 }
\hline
\end{tabular}
%}
\end{table}

\end{document}